\documentclass[letterpaper, 10 pt, conference]{ieeeconf}  

\IEEEoverridecommandlockouts                              

\usepackage{graphicx}
\usepackage{amsmath}
\usepackage{algorithm}
\usepackage{algorithmic}
\usepackage{booktabs}
\usepackage{multirow}

\title{\LARGE \bf
MAML MOT: Multiple Object Tracking based on Meta-Learning
}

\author{Jiayi Chen$^{1,2}$ and Chunhua Deng$^{1,2\dagger}$
\thanks{This work was supported by the Key R\&D projects in Hubei Province under Grant 2023BAB071 and College Students' Innovation and Entrepreneurship Training Program in Hubei Province (No. S202310488126).}
\thanks{$^{1}$School of Computer Science and Technology, Wuhan University of Science and Technology, Wuhan, 430065, China}%
\thanks{$^{2}$Hubei Province Key Laboratory of Intelligent Information Processing and Real-time Industrial System, Wuhan, 430065, China}%
\thanks{$^{\dagger}$Corresponding author(dchzx@wust.edu.cn)}%
}

\begin{document}

\maketitle
\thispagestyle{empty}
\pagestyle{empty}

\begin{abstract}

With the advancement of video analysis technology, the multi-object tracking (MOT) problem in complex scenes involving pedestrians is gaining increasing importance. This challenge primarily involves two key tasks: pedestrian detection and re-identification. While significant progress has been achieved in pedestrian detection tasks in recent years, enhancing the effectiveness of re-identification tasks remains a persistent challenge. This difficulty arises from the large total number of pedestrian samples in multi-object tracking datasets and the scarcity of individual instance samples. Motivated by recent rapid advancements in meta-learning techniques, we introduce MAML MOT, a meta-learning-based training approach for multi-object tracking. This approach leverages the rapid learning capability of meta-learning to tackle the issue of sample scarcity in pedestrian re-identification tasks, aiming to improve the model's generalization performance and robustness. Experimental results demonstrate that the proposed method achieves high accuracy on mainstream datasets in the MOT Challenge. This offers new perspectives and solutions for research in the field of pedestrian multi-object tracking.

\end{abstract}

\section{INTRODUCTION}

With the continuous advancement of video analysis technology, Multiple Object Tracking (MOT)\cite{ciaparrone2020deep} in complex scenes has garnered significant research attention. The primary objective of MOT is to accurately position, label, and predict the trajectories of multiple pedestrians by analyzing their appearance and motion information in video or image sequences. This process primarily includes two essential components: pedestrian detection and re-identification (Re-ID), both of which collectively influence the efficacy of multiple object tracking.

The rapid development of deep learning has brought theoretical and technological innovations to multi-object tracking. For example, DeepSort\cite{wojke2017simple} and CDA-DDAL\cite{bae2017confidence} effectively associate pedestrian appearance features with detection results through deep learning, achieving excellent multi-object tracking performance. RAN\cite{fang2018recurrent} and AMIR\cite{sadeghian2017tracking} use methods of autoregression and Long Short-Term Memory (LSTM) to predict pedestrian motion and appearance features. In addressing the issue of partial occlusion in multi-object tracking, STAM-MOT\cite{chu2017online} employs a spatiotemporal attention mechanism, cleverly tackling this challenge. The end-to-end deep neural network designed by RNN-LSTM\cite{milan2017online} can effectively handle the relationship between tracking and detection, achieving state updates, trajectory initialization, and termination. Furthermore, MHT-DAM\cite{kim2015multiple} and MHT-bLSTM\cite{kim2018multi} based on the Multiple Hypothesis Tracking (MHT) framework, strengthen the learning and extraction of target appearance features through convolutional neural networks and bilinear LSTM networks. These technological innovations collectively drive the development of the field of multi-object tracking.

\begin{figure}[h] 
      \centering
      \includegraphics[width=\linewidth]{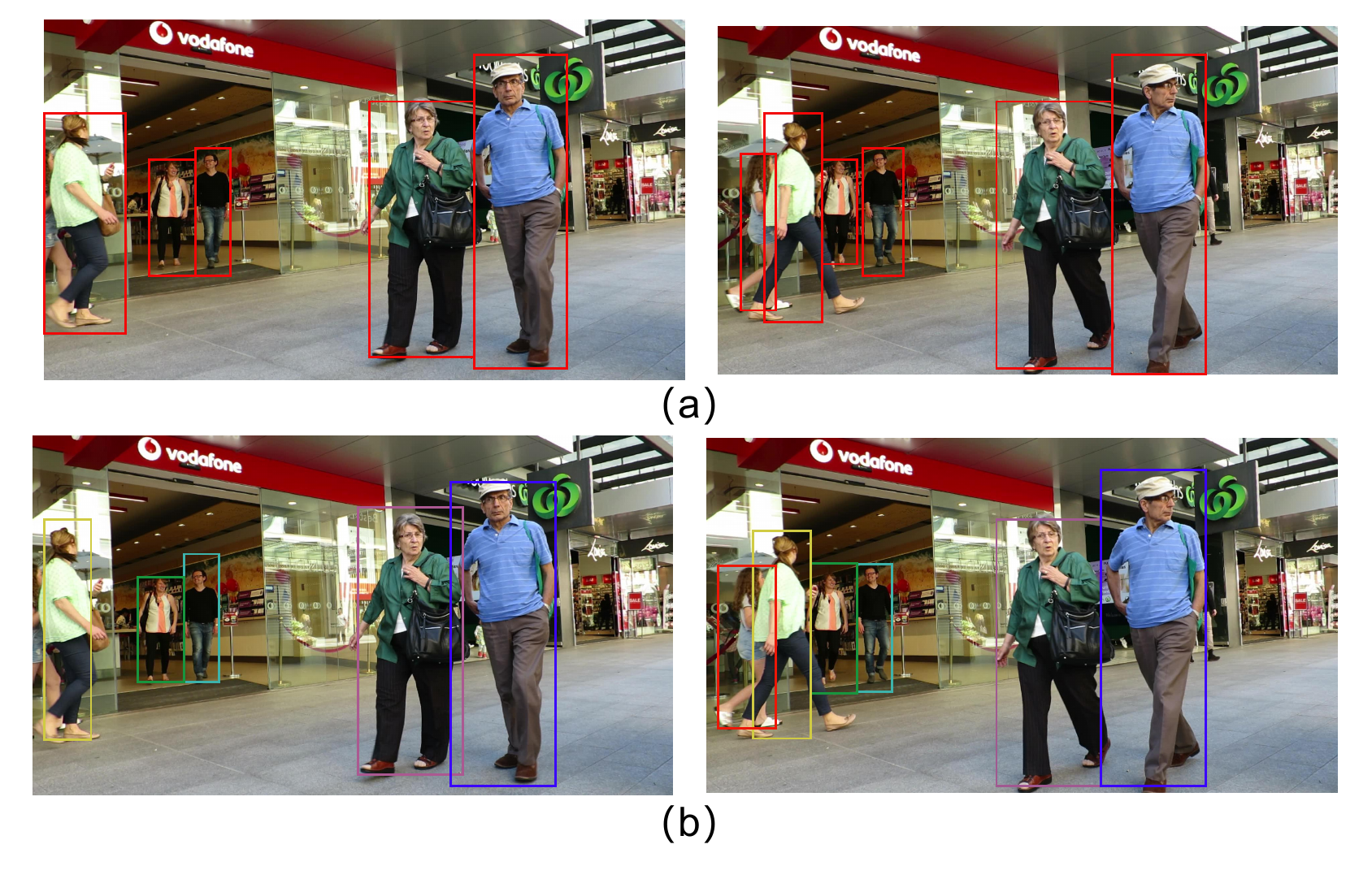}
      \vspace{-2em}
      \caption{(a) The red box indicates the pedestrian detected during the   pedestrian detection task. (b) Boxes of the same color represent the same pedestrian ID, while boxes of different colors represent different pedestrian IDs.}
      \label{fig:1}
\end{figure}

Although these technologies have been successful, supervised deep learning methods typically require a large amount of annotated data for training. While the total number of pedestrian samples in multi-object tracking datasets is relatively high, there is a long-tail effect in reality, leading to a scarcity of samples for some pedestrian categories. This data imbalance is particularly prominent in Re-ID classification tasks. Pedestrian detection tasks often perform well due to an adequate number of samples, while in Re-ID classification tasks, the limited number of samples for each pedestrian category usually results in poor model generalization. Fig.~\ref{fig:1} reveals the difference in sample numbers between pedestrian detection tasks and Re-ID classification tasks by showing two frames from a video. In the Fig.~\ref{fig:1}, (a) illustrates that in pedestrian detection, almost all pedestrians that appear can be used as training samples, while (b) shows that in Re-ID classification tasks, many pedestrians have only one or two samples due to classifying based on pedestrian identity IDs.


\begin{figure*}[ht]
      \centering
      \includegraphics[width=\linewidth]{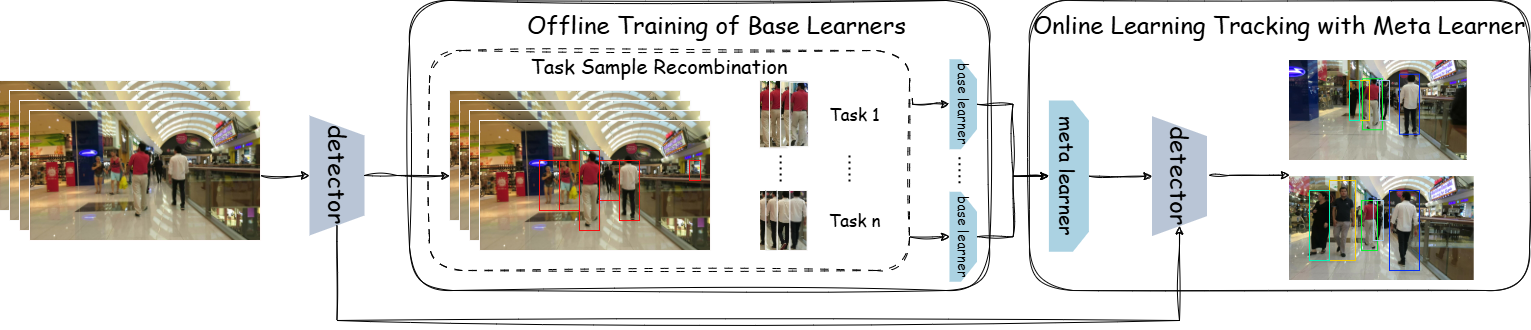}
      \caption{Overview of MAML MOT.}
      \label{fig:2}
\end{figure*}

To address this issue, commonly used data augmentation methods may increase the complexity of model training. A more effective approach to tackle this challenge is through few-shot learning. The Re-ID classification task is similar to multi-object classification tasks, with a small number of samples for each task. Inspired by human rapid learning, models should have the ability to learn across multiple tasks, not limited to just one task\cite{fei2006one}\cite{lampert2013attribute}\cite{santoro2016meta}\cite{park2018meta}. This aligns with the concept of meta-learning, which states that machine learning should have the ability to learn.

Among many meta-learning methods, Model-Agnostic Meta Learning (MAML)\cite{finn2017model} has attracted attention for its advantages in few-shot learning. MAML is characterized by its independence from specific model architectures, enabling rapid learning across various tasks and quick adaptation to new tasks with minimal data requirements. This capability enhances the generalization performance of multi-object tracking models. The core idea of MAML is to iteratively update model parameters during training, allowing the model to quickly adapt with minimal gradient updates when presented with new tasks.

Considering the few-shot learning characteristics of Re-ID classification tasks in multi-object tracking, this research proposes a multi-object tracking method based on MAML, namely MAML MOT. This method consists of two parts: offline base learner training based on MAML and online learning tracking strategy based on meta-learning. Experimental results show that this method achieves state-of-the-art performance on mainstream datasets in the MOT Challenge. Finally, the principal contributions of this research are as follows:

\begin{enumerate}
    \item the concept of meta-learning is introduced to address the few-shot learning problem in multi-object tracking.
    \item a meta-learning-based multi-object tracking method, MAML MOT, is proposed, which leverages the rapid learning ability of meta-learning to address the few-shot learning characteristics of Re-ID classification tasks. This approach aims to enhance the model's generalization and robustness, avoiding the issue of local optima
\end{enumerate}

\section{MAML MOT}

\subsection{OVERVIEW}

Multi-object tracking involves two subtasks: pedestrian detection and re-identification (Re-ID). To address the issues of limited training samples and slow training speed in the Re-ID task, we treat it as classification tasks for multiple different objects and proposes a multi-object tracking method called MAML MOT based on the MAML meta-learning algorithm.

MAML (Model-Agnostic Meta-Learning) aims to enable models to adapt to new tasks by quickly learning across multiple tasks. Its core idea is to iteratively update model parameters during training, allowing the model to rapidly adapt to new tasks with minimal gradient updates. Currently, it is mainly applied to tasks such as image classification. MAML does not require a large amount of training data but instead learns quickly across multiple similar tasks. This aligns with the characteristics of the multi-object tracking Re-ID task, which involves a variety of training types but limited training samples, hence we introduce it into the multi-object tracking task.

The MAML MOT framework comprises two primary components: offline training of base learners utilizing MAML and online learning tracking employing meta-learners. The initial phase is dedicated to identifying an appropriate initial value range with minimal data, focusing on initial parameters that are more task-sensitive. This adjustment alters the gradient descent direction, facilitating rapid model fitting on a constrained dataset. Subsequently, the second phase leverages the trained meta-learner for swift online acquisition of new tasks. The comprehensive framework is illustrated in Fig.~\ref{fig:2}.

\subsection{Offline training method of base learners based on MAML}

The objective of the offline training procedure for base learners is to develop the meta-learner, which is specifically divided into two stages. Firstly, pedestrian appearance features are extracted based on the pedestrian detection task, and training samples are restructured to construct a training set with tasks as the fundamental unit. Secondly, the hierarchical optimization method of MAML is used to train the meta-learner formed by the combination of base learners.

In general, within deep learning-based supervised learning, the training unit is a pair of data samples, which includes the original training data and corresponding label information. The network parameters are continuously updated under the supervision of labels. However, meta-learning is different from common supervised learning; its basic training unit is tasks.

\begin{figure}[h] 
      \centering
      \includegraphics[width=\linewidth]{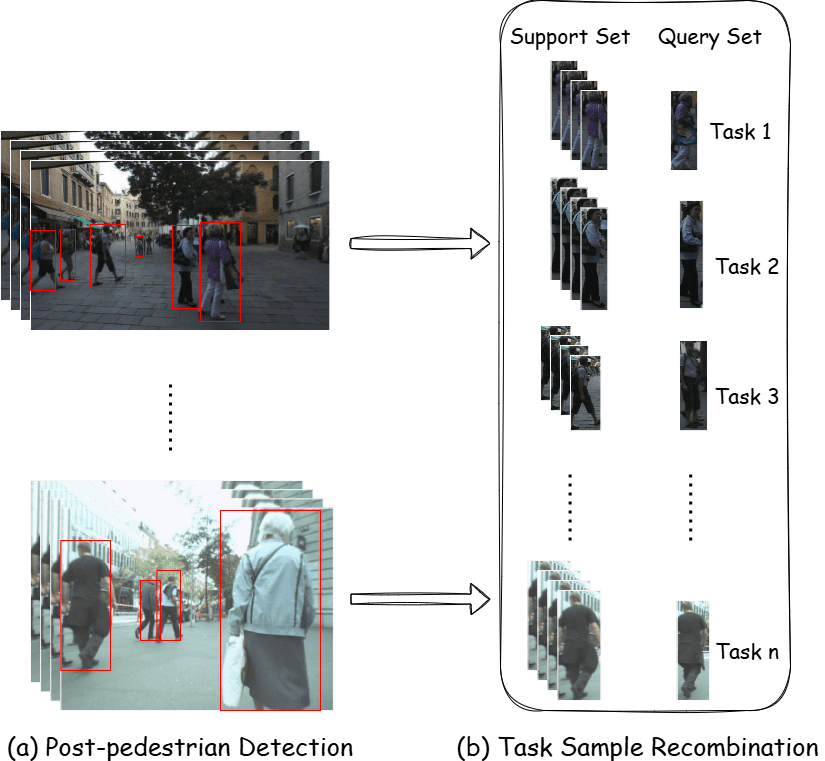}
      \caption{Multi-object tracking task sample reorganization.}
      \label{fig:3}
\end{figure}

In this work, considering the characteristics of multi-object tracking datasets, we combine pedestrian samples with the same appearance features into one task and divide the data of each subtask into support sets and query sets. The sample recombination method based on pedestrian appearance features is shown in Fig.\ref{fig:3}. In the Fig.\ref{fig:3}, (a) shows the detected pedestrians in the pedestrian detection task, (b) each row represents a task, with the left k images in each task as the support set and the right image as the query set, all containing images and corresponding label information.

Task organization is based on subsequences, where the support and query sets of the same task come from different frames of the same video sequence to ensure that the data used for single-step learning gradient updates and error calculations are consistent, while there are no requirements for different task data. Equation (1) presents the formal expression of the task sample recombination method, where $T$ represents the task, $S$ represents the support set, $Q$ represents the query set, and $P$ represents the image:

\begin{equation}
    \begin{gathered}
        T=\left\{T_1,T_2,\ldots,T_n\right\} , \\
        T_i=\left(S_i,Q_i\right)\ \ \ \ \ \ i=1,2,...,n\ , \\
        S_i=\left\{P_1,P_2,\ldots,P_k\right\}, Q_i=\left\{P_{k+1}\right\}.
    \end{gathered}
\end{equation}

\begin{figure}[h] 
      \centering
      \includegraphics[width=\linewidth]{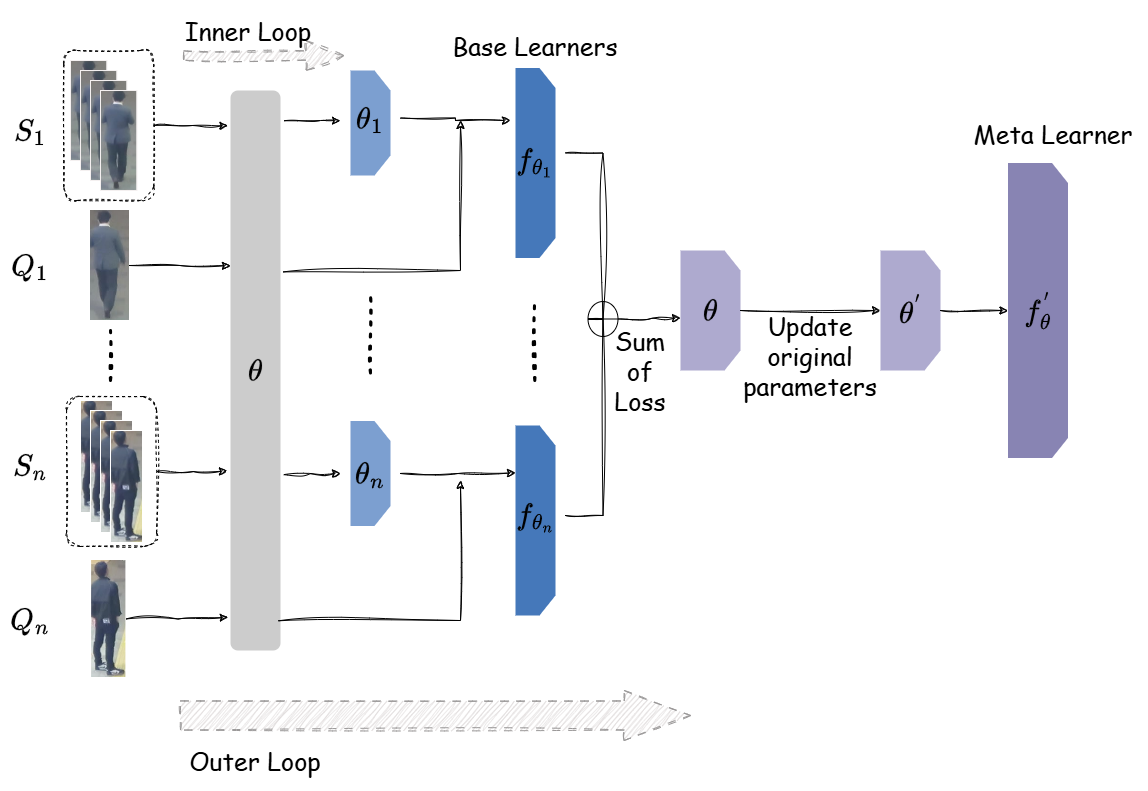}
      \caption{Meta-learning hierarchical optimization process.}
      \label{fig:4}
\end{figure}

The MAML hierarchical optimization method does not directly update parameters using data and labels in a single step, unlike the general training method. It is a hierarchical optimization process. However, it shares a commonality with the common training process in that both use gradient descent-based methods to update parameters. Fig.\ref{fig:4} illustrates the hierarchical optimization process based on MAML in this study, specifically divided into inner optimization and outer optimization, where $\left(S_i,Q_i\right)$ represents the combination of the i-th task's (support set, query set). Algorithm \ref{alg:1} summarizes the hierarchical optimization process.

The goal of inner optimization is to train a base learner and its temporary parameters for a set of tasks. Inner optimization calculates multiple times based on different $\left(S_i,Q_i\right)$ pairs. In order not to affect the outer network parameters $\theta$ during inner optimization, it is necessary to first backup $\theta$, as shown in the gray area in the figure. The initial parameters for each inner optimization are $\theta$. We first update $\theta$ with the support set once, obtaining temporary parameters ${\theta_i}$. Then, using the query set ${Q_i}$, it calculates the loss value ${L_i}$ on the parameters ${\theta_i}$. At this point, the temporary parameters ${\theta_i}$ can be released, and all ${L_i}$ are summed to be used for outer optimization updates.

The goal of outer optimization is to synthesize a meta-learner from multiple base learners trained for inner optimization. Specifically, outer optimization will use the previously backed-up parameters $\theta$. We calculate the gradient on $\theta$ using the sum of $L_i$, and then update the parameters $\theta$ to ${\theta^\prime}$ using an outer loop learning rate. At this point, one iteration of outer optimization is completed, and the next round of outer loop optimization begins.

\begin{algorithm}[H]
\caption{MAML hierarchical optimization.}
\label{alg:1}
\textbf{Input:} distribution over tasks $P\left(T\right)$, step size hyper-parameters $\alpha,\beta$ \\
\textbf{Output:} trained network model parameters $\theta$
\textbf{begin}
\begin{algorithmic}[1]
\STATE randomly initialize $\theta$
\STATE // outer loop
\WHILE{not done}
    \STATE Sample batch of tasks $T_i\left(S_i,Q_i\right) \sim P(T)$
    \STATE // inner loop
    \FOR{all $T_i$}
         \STATE Evaluate ${\ \nabla}_\theta L_{T_i}\left(f_\theta\right)$ with respect to $K$ examples in $S_i$
	    \STATE compute adapted parameters with gradient descent $\theta_i^\prime=\theta-\alpha\nabla_\theta L_{T_i}\left(f_\theta\right)$
    \ENDFOR
    \STATE Update $\theta \leftarrow \theta-\beta\nabla_\theta\sum_{T_i\sim P(T)} L_{T_i(f\theta_i^\prime)}$
\ENDWHILE
\end{algorithmic}
\textbf{end}
\end{algorithm}

The basic idea of the offline training method we propose is to train a set of cross-task base models, perform a small amount of gradient descent on the initialized parameters, and adapt to new tasks with only a small amount of data. From the perspective of gradients, by training on a large number of tasks, the gradient information learned from different tasks is superimposed into a fused gradient for updating model parameters, making the model sensitive to maximizing the loss function of new tasks.

During the inner optimization stage, the base learner and its parameters are represented by $f_\theta$ and $\theta$. When $f_\theta$ adapts to a new task $T_i$, its parameters transition from $\theta$ to $\theta^\prime$.  MAML employs the gradient descent method on task $T_i$ to update the parameter vector $\theta^\prime$, as illustrated in equation (2):

\begin{equation}
        \theta_i^\prime=\theta-\alpha\nabla_\theta L_{T_i}\left(f_\theta\right).
\end{equation}

Equation (2) shows the process of updating parameters using one gradient descent step, where $\alpha$ is a hyperparameter representing the learning rate, $L_{T_i}(f_\theta)$ denotes the loss function of the base learner on task $T_i$, and $\nabla_\theta$ represents the gradient of the loss value with respect to the current parameter $\theta$.

The objective function of the base learner is shown in equation (3), where $\theta$ and $\theta^\prime$ represent the parameters of the base learner before and after the update, respectively:

\begin{equation}
    \min_{\theta_{i}^{\prime}}L_{T_{i}}\left(f_{\theta_{i}^{\prime}}\right)=L_{T_{i}}\left(f_{\theta-\alpha\nabla_{\theta}L_{i}\left(f_{\theta}\right)}\right).
\end{equation}

During the outer optimization stage, given the sampling distribution $P(T)$, the base learner feeds back $\theta_i^\prime$ to the meta-learner. At this point, the objective function of the meta-learner is as shown in equation (4):

\begin{equation}
        \min_{\theta}\sum_{T_i\sim P(T)}{L_{T_i}(f_{\theta_i^\prime})}=\sum_{T_i\sim P(T)}{L_{T_i}(f_{\theta-\alpha\nabla_\theta L_i(f_\theta)})}.
\end{equation}

Simultaneously, the initial parameters $\theta$ of the meta-learner are updated through gradient descent, as shown in equation (5), where $\beta$ is a hyperparameter representing the learning rate for the second update of the model:

\begin{equation}
        \theta\gets\theta-\beta\nabla_\theta\sum_{T_i\sim P(T)} L_{T_i}(f_{\theta_i^\prime}).
\end{equation}

Ultimately, an initial model parameter $\theta$ is obtained, which possesses task-sensitive characteristics. Subsequent tasks can be adjusted online based on this initial model to adapt to the requirements of different tasks.

\subsection{Online learning tracking strategy of meta-learner}

We consider the online environment as the testing environment, where each frame of the video sequence is not included in the training set of offline training. Each group of pedestrians with the same appearance features appearing in each frame of this sequence is regarded as a new task.

After obtaining a cross-task Re-ID classification meta-learner through offline training, we first initialize the meta-learner $f_{\theta_{new}}$ and the parameters $\theta_{new}$ using the first frame image of each new task $T_{new}$, as shown in equations (6) and (7), where  ${\gamma}_i$ represents the similarity between the new task $T_{new}$ and the training task $T_i$, and $k$ denotes the number of support set samples in training task $T_i$:

\begin{equation}
    \begin{gathered}
        \begin{aligned}
            L_{T_{new}}(f_{\theta_{new}})&\approx\frac{1}{\lambda}\sum_{T_i\sim P(T)}{\gamma_iL_{T_i}(f_{\theta_i^\prime})},\\ 
            \lambda&=\sum_{T_i\sim P\left(T\right)}\gamma_{i\ },\\
        \end{aligned}
        \\
        \begin{aligned}
            \gamma_i&=Similarity\left(T_{new},T_i\right) \\
            &=\frac{\sum_{j=1}^{k}\left(T_{new_1}\times T_{i_j}\right)}{\sqrt{\sum_{j=1}^{k}T_{new_1}^2}\times\sqrt{\sum_{j=1}^{k}T_{i_j}^2}}.
        \end{aligned}
    \end{gathered}
\end{equation}

\begin{equation}
        \theta_{new}=\frac{1}{\lambda}\sum_{T_i\sim P\left(T\right)}{\gamma_i\theta_i}.
\end{equation}

Next, we perform the same operation on subsequent frames, calculating Re-ID classification using the cosine similarity loss function.

It is worth noting that as mentioned in section B, the meta-learner is trained only on the last layer of the backbone network. Therefore, when using the meta-learner for classification in this section, parameters are shared, and only the parameters of the last layer of the meta-learner are computed, significantly improving the speed of online learning.

\section{EXPERIMENT}

\subsection{Dataset and Evaluation Metrics} 

We use mainstream pedestrian tracking datasets, namely the MOT Challenge dataset, including MOT16\cite{milan2016mot16}, MOT17\cite{milan2016mot16}, and MOT20\cite{dendorfer2020mot20}.

\begin{itemize}
\item MOT16: Contains $7$ training sequences and test sequences, with $5316$ and $5919$ images in the training and test sets, respectively. The training set includes $112K$ bounding boxes and $0.5K$ IDs for training. 

\item MOT17: The sequences are identical to MOT16, but MOT17 provides pedestrian detection results from three public detectors, allowing tracking models to directly reuse these detection results, eliminating the impact of object detection on the overall tracking results for fair comparison in public competitions. Additionally, MOT17 has updated some annotations in the dataset. 

\item MOT20: Consists of 8 video segments from three different scenes, with 8931 frames in the training set and 4479 frames in the test set. It includes videos with higher crowd density captured in unrestricted environments, some shot at night with significant occlusions between pedestrians, posing high challenges.
\end{itemize}

We evaluate our method on the corresponding test set and submit it to the MOT Challenge server. It is worth noting that all results are directly obtained from the official MOT Challenge evaluation server.

We use CLEAR MOT Metrics\cite{bernardin2008evaluating} (such as MOTA, IDs) and IDF1\cite{ristani2016performance}  as evaluation metrics. Specifically, the Multiple Object Tracking Accuracy (MOTA) measures the accuracy of the model in tracking targets by calculating false positives, false negatives, and mismatches in tracking all targets. A higher MOTA value indicates better performance, as shown in equation (8), where False Positive (FP) and False Negatives (FN) respectively represent the number of negative samples predicted as positive and positive samples predicted as negative. IDs indicate the number of ID switches in trajectories, and GT represents the total number of ground truth values:

\begin{equation}
        MOTA=1-\frac{\sum{(FN+FP+{ID}_s)}}{GT}.
\end{equation}

The IDF1 metric is the proportion of correctly associated targets to the sum of all ground truth targets and detected targets. It focuses more on the trade-off between ID precision and ID recall. The TP, FP, and FN metrics all consider ID information, so IDF1 can reflect the ability to maintain target identity, as shown in equation (9), where Identification Precision (IDP) refers to the precision of target ID identification, while Identification Recall (IDR) denotes the recall of target ID identification. IDTP is the number of true positive IDs predicted as positive, IDFP is the number of false positive IDs predicted as positive, and IDFN is the number of false negative IDs predicted as negative:

\begin{equation}
    \begin{gathered}
        \begin{aligned}
            IDF1&=\frac{2}{\frac{1}{IDP}+\frac{1}{IDR}} \\
            &=\frac{2IDTP}{2IDTP+IDFP+IDFN}.
        \end{aligned}
    \end{gathered}
\end{equation}

\subsection{Implementation details}

We train the model on the training sets of three benchmark datasets, including MOT16, MOT17, and MOT20. The model is then tested on the corresponding test sets of each training set, and the predicted results are submitted to a public server to obtain the final experimental results. For fair comparison, the experimental results of other models cited in this study are all from the ranked list on the public server.

In the study, we use the CLEAR MOT metric and IDF1 to validate tracking accuracy. During training, common data augmentation methods such as rotation, scaling, and affine transformation are applied to process images. The images are then resized to $1088\times608$, with a downsampling rate of $4$ for the network, resulting in feature maps of size $272\times152$.

During the meta-learning pre-training process, the inner optimization learning rate is set to $5\times{10}^{-3}$, the outer optimization learning rate is set to ${10}^{-4}$, and training is conducted for $60$ epochs. In the subsequent model training process, we use the Adam optimizer to train the network for $30$ epochs, with an initial learning rate of ${10}^{-4}$, reduced to ${10}^{-5}$ at the 20th epoch. The batch size is set to $12$, and training is completed on 2 RTX2080 Ti GPUs.

\subsection{Analysis of Results}

Our method is trained on the training sets of three benchmarks, obtained results on the corresponding test sets, and compared them with the SOTA methods. Table \ref{tab:1} shows the experimental results on the MOT16, MOT17, and MOT20 test sets. Bold indicates the best value for the corresponding metric, and underline indicates the second-best value.

\begin{table}[h]
\caption{Results on the MOT16, MOT17, and MOT20 test sets, as well as the results of the comparison methods.}
\label{tab:1}
    \centering
    \begin{tabular}{ccccc}
        \toprule
        \textbf{dataset} & \textbf{methods} & $\boldsymbol{MOTA}$ & $\boldsymbol{IDF1}$ & $\boldsymbol{ID_S}$ \\
        \midrule
            \multirow{6}{*}{\textbf{MOT16}} & QDTrackX\cite{fischer2023qdtrack} & - & - & -  \\
                              & TLR\cite{wang2021multiple} & 76.6 & 74.3 & 979 \\
                              & SGT\_mot\cite{hyun2023detection} & 76.8 & 73.5 & 1276 \\
                              & CountingMOT\cite{ren2022countingmot} & 77.6 & 75.2 & 1087 \\
                              & Bot-SORT\cite{aharon2022bot} & \underline{78.6} & \underline{79.1} & \underline{662} \\
                              & MAML MOT(Ours) & \textbf{79.7} & \textbf{80.2} & \textbf{469} \\
        \midrule
            \multirow{6}{*}{\textbf{MOT16}} & QDTrackX & 78.7 & \underline{77.5} & \underline{1935}  \\
                              & TLR & 76.5 & 73.6 & 3369 \\
                              & SGT\_mot & 76.3 & 72.4 & 4578 \\
                              & CountingMOT & 78.0 & 74.8 & 3453 \\
                              & Bot-SORT & \underline{79.4} & \textbf{79.4} & 2643 \\
                              & MAML MOT(Ours) & \textbf{79.9} & 77.3 & \textbf{1890} \\
        \midrule
            \multirow{6}{*}{\textbf{MOT16}} & QDTrackX & 74.7 & 73.8 & \textbf{1042}  \\
                              & TLR & - & - & - \\
                              & SGT\_mot & 72.8 & 70.5 & 2649 \\
                              & CountingMOT & 70.2 & 72.4 & 2795 \\
                              & Bot-SORT & \underline{75.1} & \underline{75.9} & \underline{1213} \\
                              & MAML MOT(Ours) & \textbf{76.2} & \textbf{76.6} & 1358 \\        
        \bottomrule
    \end{tabular}
\end{table}

From the table above, it can be seen that our proposed MAML MOT method shows a slight improvement in the MOTA metric compared to directly training methods, a significant decrease in IDs, indicating that the network performs well in target detection tasks while reducing the ID switch rate. Our method shows an improvement or nearly the same performance in the IDF1 metric compared to other methods, indicating an enhancement in trajectory association quality. This suggests that the meta-learning optimization method has promoted the learning of Re-ID tasks to some extent, providing significant assistance in maintaining target identity.

\section{CONCLUSIONS}

This study addresses the challenge of limited generalization capability in Re-ID classification tasks, stemming from the intricate nature of pedestrian multi-object tracking environments, the variety of pedestrian categories, and the scarcity of instance samples. It incorporates the MAML meta-learning approach into multi-object tracking endeavors, proposing a meta-learning-driven multi-object tracking methodology, MAML MOT. Specifically, the methodology comprises an offline training approach for the base learner utilizing MAML and an online learning tracking strategy for the meta-learner. The former optimally leverages the attributes of multi-object tracking datasets with diverse sample types and limited instances for meta-learning task segmentation and hierarchical optimization. The latter entails swift online learning of novel tasks based on the meta-learner derived from offline training, striking a balance between learning from limited samples and enhancing the model's generalization capacity. The tracking outcomes on the MOT Challenge public dataset affirm the efficacy of our approach.

\bibliographystyle{IEEEtran}
\bibliography{root}

\addtolength{\textheight}{-12cm}   










\end{document}